# Self-Organizing Machine Translation :
# Example-Driven Induction of Transfer Functions

Patrick Juola

# University of Colorado at Boulder

# Self-Organizing Machine Translation :
# Example-Driven Induction of Transfer Functions

Patrick Juola

May 1994

**Abstract**

*Come, let us go down and there make such a babble of their language that they will not understand another's speech.* – Genesis 11:7

With the advent of faster computers, the notion of doing machine translation from a huge stored database of translation examples is no longer unreasonable. This paper describes an attempt to merge the Example-Based Machine Translation (EBMT) approach with psycholinguistic principles. A new formalism for context-free grammars, called *marker-normal form*, is demonstrated and used to describe language data in a way compatible with psycholinguistic theories. By embedding this formalism in a standard multivariate optimization framework, a system can be built that infers correct transfer functions for a set of bilingual sentence pairs and then uses those functions to translate novel sentences.

The validity of this line of reasoning has been tested in the development of a system called METLA-1. This system has been used to infer English→French and English→Urdu transfer functions from small corpora. The results of those experiments are examined, both in engineering terms as well as in more linguistic terms. In general, the results of these experiments were psychologically and linguistically well-grounded while still achieving a respectable level of success when compared against a similar prototype using Hidden Markov Models.

# 1 Introduction

## 1.1 Problem Statement

This report describes an approach to automatic translation of natural languages that accepts a bilingual sentence database and infers a set of transfer functions for converting sentences from one language into the other. By focusing on some results from experimental and developmental psycholinguistics, as expressed in a computational formalism developed below, one hopes to bring some novel order to the very chaotic world of human languages. Specifically, some of the same properties that make human languages learnable by human infants should be able to provide syntactic and semantic cues that an appropriately programmed computer can use for a variety of NLP problems.

Every child who has learned her native language can identify with great accuracy the grammatical utterances in that language. On the other hand, this task is nearly impossible for computers. Gold[21] showed that under reasonable assumptions (that people only hear positive examples of the language and that any finite string, in theory, *could* be a sentence in some language) this task is formally impossible for a computer to solve. Specifically, it is impossible to avoid overgeneralization. Once a sentence has been assumed to be in the language, there is no way to accept or act upon information to demonstrate that it is *not* actually a grammatical sentence. For all the algorithm knows, it might simply be a sentence that it hasn't yet heard.

Several solutions have been proposed to avoid the despairing abyss of Gold's theorem. For instance, it has been pointed out that grammaticality judgements, despite their apparent simplicity, are actually an unrepresentative task. For most natural language applications, it is more important to determine what a speaker means even in the event of an ungrammatical sentence than to reject it out of hand. Similarly, sentences can be more usefully categorized as appropriate/inappropriate; the sentence "I like pizza" is perfectly grammatical, but would be wrong as a response to the question "What was your grade on the final?"

Machine translation is a suitable application to test these ideas for a variety of reasons. Because the outputs of a translation system are more complex than simple yes/no decisions, more information is available to limit and direct the inference task. Second, translation is a somewhat nebulous task, as there may be many sentences in the target language which can express the idea of the source sentence. More importantly, there is a nearly continuous spectrum of near-correctness, and so the notion of accuracy can be more finely developed. A system is successful not when it has learned "the translation," but when it is close enough to perfect that it is capable of expressing the ideas present in the source text in the target language to a high enough degree of satisfaction. Third, and most importantly, the system under development uses constraints developed from psycholinguistic experiments to develop its functions, which should result in functions that can be symbolically and linguistically studied, modified, and maintained. In contrast to the typical development process of symbolic translation systems, though, the system is completely automated and can run without the explicit development by hand of a dictionary, thesaurus, or analysis engine. This final aspect, if successful, will represent a major improvement over the current state of the art in Machine Translation.

## 1.2 Organization

The rest of this report is organized as follows : Section 2 describes some of the relevant work that has been performed, both on the problem of machine translation, and computational and psycholinguistic approaches to language acquisition. Section 3 describes the formal basis for this psycholinguistic approach to machine translation, including the major mathematical result of the paper, the existence of a "marker-normal form" for the description of context-free grammars. The METLA-1 system as currently implemented is described in detail in section 4, along with the results of several experiments to determine the the strengths and weaknesses both of the formalism and its implementation, presented as section 5. Finally, section 6 suggests some needed improvements to be included in future versions, followed by some conclusions, acknowledgements, and a bibliography.

## 2 Background

### 2.1 Machine Translation

Translation is a hard process. A typical translation task requires a skilled bilingual with expert training in the translation process itself and in-depth knowledge of the document's field. Even so, it often requires the additional assistance of outside experts for their additional knowledge of the field and the expected audience.

The morass is no shallower when computers are allowed into the translation process. Instead of being able to use documentary cues, and to translate on a case-by-case basis, computer scientists must predict beforehand the types of situations that will arise and armor their programs against them. A typical MT system can take man-years of effort from skilled programmers, knowledge engineers, linguists, and translators, and will cost as much as a conventional rule-based expert system. Even then, most of machine translation's success stories are limited systems, based on small domains (such as weather reports, aviation, conference registration, and so on). In these areas, the vocabulary, grammar, and meanings are small, tightly focused, and very predictable. Even after legions of scientists have devoted their efforts to it, MT has not yet advanced far beyond the realm of toy problems.

The major bottleneck, of course, is the acquisition of knowledge and the processing into a form that a computer program can recognize and work with. Each word in a language, for example, may have several translations in any other language, to be used in "appropriate" contexts, and the definition of appropriate is sometimes an entire linguistics dissertation. For some early systems (such as the GAT/SYSTRAN project), the research approach has been to build (over the course of decades) a tremendous list of word, phrases, and idioms and to translate sentences word-for-word and phrase-for-phrase without any real structural representation. A more popular approach has been to try to develop an automatic method for extracting the structure and meaning of the source document and converting it, whether by transformation or retelling, into the target language. This obviously requires a significant linguistic understanding of both the source and target languages to tease apart the necessary constraints and interdependencies—for example, the English verb "to wear" corresponds to five Japanese verbs, depending upon the object worn. This information is painstakingly gleaned from the skilled bilinguals and carefully translated into computer code to represent each new construct or constraint included.

However, all the information a linguist or translation team could require is available in the huge body of translated text that already exists—if it could be appropriately extracted and put to use. As an example, the Canadian parliamentary proceedings (named the Hansard corpus, after the publisher) are available in both English and French, and constitute a huge corpus of grammatical information about the structure and relationship of those languages.

Brown et al.[6, 7] tried to use the Hansard corpus to develop a completely automated MT system for French and English. They attempted to describe language as a Markov process, and the translation process as a mapping between Markov chains of words. By using automated techniques, they attempted to refine the probability distributions so that the "expected sentence" corresponded to the translated sentence over as much of the training data as possible. The results of this were surprisingly good, given what one might expect from such a linguistically implausible formulation. On the other hand, trying to actually understand the resulting system (for instance, to explain or correct systematic errors) would be a nightmare. An ideal translation system should have a better representation of the structure of the relevant languages, if only to make system maintenance easier.

Nagao[37] proposed exactly this when he invented the term "Example-Based Translation." Instead of building explicit translation rules and vocabulary, a system could be built that uses a database of hundreds or thousands of sentences and phrases with their corresponding translation. These could provide coverage *in context* for the appropriate parts of input sentences. The system, then, would break input sentences apart into appropriately-sized chunks, search the database for the most similar example phrases, and build the translated sentence by skillfully-developed conglomeration of similar phrases.

EBMT, by Sumita et al.[47], is an early example of a Japanese→English translation system built along these lines. It translates phrases of the form $noun_1$ *NO* $noun_2$, where '*NO*' is the general partitive adposition[1]. The standard translation of '*NO*' is 'of', but it can also be translated as 'in' (as in "the conference in Kyoto"), ''s' ("the teacher's pencil"), 'by', and many other (typically prepositional) translations. Sumita et al. incorporated a commercial thesaurus into their system, and devised a notion of semantic distance to measure the nouns in the source sentence against the various examples in the database. The example closest, in semantic distance, to the source sentence was chosen as the basis of translation, and then the target sentence was produced and structured according to the example.

From one point of view, EBMT is simply a huge dictionary entry for the single word '*NO*', with the appropriate contextual cues. However, the work done to build the translation database and the thesaurus can be easily reused in a larger system, and the general method can greatly reduce the knowledge acquisition bottleneck. By relying on translations in context, the amount of work that need be done per lexical entry is much smaller, and the system is much more easily adaptable to different styles, situations, and fields. This approach has been taken to much larger systems; [40, 39] lists some example systems, where the translation mechanism described above is attached to a general parser and an adjustment mechanism to combine the translations of the parsed fragments. The use of translation examples greatly simplifies the task of constructing a dictionary, but the development of a thesaurus, parser, analyzer, and adjuster still require lots of language-specific skilled labor. Although EBT-based approaches reduce the knowledge-acquisition bottleneck, they do not eliminate it. To do that would require a totally self-organizing system (after [6]) that produces linguistic descriptions which can be understood, adjusted, and modified by non-experts.

---

[1] Some languages, such as Japanese, have their prepositions follow nouns instead of precede them. ("*in* Tokyo" becomes "Tokyo *in*") "Adposition" is a general term for this type of word, without regard to position.

## 2.2 The Marker Hypothesis

The first step in developing an automatic analyzer for a novel language, of course, is to understand the nature and properties of the language. In an ideal universe, this would be a complete and accurate description of the sentences that could be said in a recognizable and usable form, such as a grammar. However, as discussed above, the extraction of grammars from example sentences is problematic. According to Gold's theorem, for a computer to identify a grammar requires either a teacher of some sort, to provide negative information, a revision (such as probabilistic learning) to the computational paradigm, or additional assumptions about the nature and structure of the language.

Are there, then, properties of natural languages that are not found or not prominent in the "mathematical" languages described by formal structures such as context-free grammars (CFGs)? Children are certainly capable of learning their native language, despite the fact that most if not all of their data consist only of positive examples of acceptable, grammatical sentences. Very rarely do children receive corrections about ungrammatical utterances, and even more rarely do they appear to attend to the few corrections they receive. So if Gold's theorem applies to humans as well as to computers, there may be some linguistic absolutes that apply to natural languages that allow children to identify the structures of the language.

This search for "linguistic universals" is the major focus of the field called linguistic typology, and it has not been without its major successes. For example, Berlin and Kay[3] identified a hierarchy in the existence of basic color terms, demonstrating that some basic colors are present in every language and that the set of optional color terms forms a well-defined sequence. Keenan and Comrie[30] identified a similar hierarchy of "accessibility to relativization," showing (for instance) that any language which allows relativization of indirect objects allows relativization of direct objects but not vice versa. In addition to these and other micro-features of the nature of human languages, Greenberg[23] identified major structural regularities among the world's languages—for example, in languages with basic word order subject–object–verb, the adpositions tend to be placed after the nouns they modify. Greenberg identified nearly fifty such universals which together describe basic structural properties of natural language in terms of a small number of parameters and simultaneously describe a large set of theoretical possibilities which do not occur. A computer program could, in theory, focus on such parameters and by determining the settings for them identify a language much more quickly than by exhaustive searching through the entire mathematical space of CFGs.

Analyzing a particular language in terms of which universals describe it is not new. For example, Chomsky[12] and his followers have proposed the notion of a "universal grammar" built into the human mind that partially limits the set of languages that exist in the world. Individual languages can be described in terms of which constraints and rules are applicable to them—e.g., English and German are not allowed to have "null subjects" that are understood but unexpressed, while Spanish and Hebrew, by contrast, allow these sorts of subjects. Dorr[16] has built this theory into a machine translation system, where universal principles are parameterized and set by hand to define the source and target languages. In some ways, this can avoid the knowledge acquisition bottleneck described above, reducing a large number of specific rules to a small set of properties. On the other hand, these properties are often controversial and difficult to elicit. Even as simple a statement as "this language has subjects before verbs" can be difficult to test against a language where the agent of a sentence, the focus of the sentence, and the topic of the sentence can be three different words. In addition, identification of syntactic or semantic categories such as "subject" is difficult without extensive (and expensive) linguistic analysis.

The work described here focuses on the computational implications of a psycholinguistic universal first expressed by Green[22] and explored and expanded by other researchers[34, 35]. This universal, termed "the Marker Hypothesis," states that natural languages are "marked" for grammar at surface level—that there exists in every language a small set of words or morphemes that appear in a very limited set of grammatical contexts and that can be said, in a sense, to signal that context. As an example of this principle, consider a basic sentence in English :

> The Boulder Faculty Assembly announced a list of ten faculty awards at its Thursday meeting, with more awards for excellence in teaching than expected.

In this sentence, taken at random from a Boulder newspaper, two noun phrases began with determiners, two with quantifiers, and one with a possessive pronoun. The set of determiners and possessive pronouns in English is very small (less than fifteen words), and the set of quantifiers is equally recognizable. Similarly, every word in this sentence ending with '-ed' is a past tense verb. The Marker Hypothesis presumes the converse of these observations, e.g. that words which end in '-ed' are very often past tense verbs, and the word 'the' usually heralds the appearance of a noun phrase. Or, more generally, that concepts and structures like these will have similar morphological or structural marking in all languages.

Proponents of the Marker Hypothesis go further, however, claiming not only that these "marker words" could signal the occurrence of particular contexts, but that they do—that marker words form an important cue to psycholinguistic processing of structure. Experiments with miniature languages have backed up this claim. When human subjects are presented with the task of learning a small artificial language from sentences in the language, they learn more accurately and faster if the artificial language has cues of the sort described above. Green[22] showed this effect in artificial languages with and without specific marker words as attested in Japanese. Morgan et al.[34] demonstrated it in languages with and without phrase-level substitutions, as of pronouns for full noun phrases. Mori and Moeser[35] examined the effect of case marking on the pseudowords of the languages. In these and other experiments, evidence confirming the Marker Hypothesis was always found.

Other evidence for the psychological utility of marker words can be found in typological evidence. The original statement of the Marker Hypothesis was based upon the typological observation that every natural language has such constructs, whether in derivational morphology or separate marker words. Even pidgins and creoles have such constructs. For example, [43] lists examples from a pidgin called Russenorsk. In this language, sentences tend to be very simple strings of words, without grammatical affectation. Even in this language, however, verbs are marked with a special '-om' marker, which presumably helps hearers of this language identify the basic concept expressed in a given utterance (and from that determine the appropriate roles of the other words in the sentence).

Other psycholinguistic evidence for such the Marker Hypothesis can be taken from child language acquisition. Constructs which are easily and readily marked (e.g., regular verbs) tend to be learned early and strongly, and may even override other irregular forms which have been learned by rote memorization. The classic child's sentence "*I goed to the store"[2] is an obvious example of this sort of overgeneralization. The child has learned that events which have already happened are described by verbs marked with the '-ed' morpheme. Slobin[44] lists dozens of psycholinguistic principles that may describe how children focus on important bits of the language to learn. Many of these (for example, "pay attention to the ends of words") are direct descriptions of phenomena the Marker Hypothesis would predict.

---
[2] An initial asterisk ('*') in front of a string is the standard linguistic notation for an ungrammatical sentence or construct.

Finally, there is psychological evidence about not only the universality of marker words and morphemes, but also about their cross-linguistic similarity. Certainly, such concepts as case marking, gender, and tense seem to be concepts found in a large variety of languages. Talmy[48] suggests that, in fact, there are certain cognitive aspects or concepts that are inherently likely to be expressed grammatically (using marker morphemes or structural cues) and others that are universally expressed lexically. For example, many languages have inflections on nouns to express the number. On the other hand, there is no known language where morphemes exist to differentiate red nouns from blue nouns. Color, then, is not a concept expressed grammatically. Similarly, languages are capable of making inflectional distinctions between the numbers '1' and 'many', or in some cases the numbers '1', '2', 'few' and 'many', but never between 'odd' and 'even', or 'prime' and 'composite'. The implication is not only that marker constructs exist, but that the semantic concepts and distinctions that they express tend to be expressed in other languages by other marker constructions.

Assuming, then, that the Marker Hypothesis is an accurate description of a useful property of natural languages, it is reasonable to use this property in an attempt to build a system that will naturally acquire the grammar of the source and target languages of interest to a machine translation system. Section 3 describes a computational formalism to do exactly that, based on the notion of acquisition of translation functions from large corpora as described above.

## 2.3 Translation and Grammatical Induction

Gold's theorem, as presented in the introduction, has not completely stopped research into grammatical induction. Despite the overall implications that context-free grammars (and even regular expressions) are formally unlearnable from positive examples alone, various researchers have modified the problem statement slightly to achieve positive results. For example, several researchers have limited the scope of the languages under study. Lucas and Damper[32], for example, have designed an algorithm capable of learning non-recursive context-free grammars (for example, languages where no sentence can be a part of another sentence, or no noun phrase can be part of a larger noun phrase) from positive examples alone.

The more psycholinguistically plausible the restricted languages are, of course, the more interest one should display in the results. Recursive productions, unfortunately, are very common, whether in lists of items ("John bought a pizza with sauce, extra cheese, onions, pepperoni, black olives,..."), or sentences that are themselves parts of other sentences ("My father told me that his boss told him that ..."). However, more plausible restrictions have been studied. Angluin[1] described a restricted set of languages called *pattern languages*. These languages are a subset of regular expressions produced by concatenation and the Kleene *plus* operator.[3] These languages, then, are in some sense self-marking, where each component marks not only itself but further appearances of the same component. Angluin showed that by focusing on this marking aspect, these languages are learnable from positive examples only.

---
[3]This operator produces, from a string, the set of all strings consisting of one or more repetitions of that string. For example, the set $\{\,a, aa, aaa, \cdots\,\}$ can be denoted more simply as $a^+$.

The unification of these and other approaches is relatively simple. Non-recursive grammars cannot produce infinite languages, and selecting the smallest (finite) language compatible with the input seen so far will converge on the correct language. Angluin's pattern languages may contain infinite languages, but has so many excluded *finite* languages that it is easy to determine when an input set determines an infinite language. Again, selecting the "smallest" compatible language will eventually converge.

A more interesting reformulation of the problem originated with the notion of probabilistic grammars. These are normal grammars, either regular or context-free, annotated with probabilities at each production rule. This allows grammars to be tuned not only to the presence and/or absence of individual strings, but to their relative frequencies in the input data. This can help solve the Gold's overgeneralization problem, since an algorithm can, in theory, recognize that a sentence occurs with much lower than predicted frequency and the grammar might have overgeneralized. In addition, probability information can improve linguistic performance on ambiguous data. For instance, the oft-cited sentence "Time flies like an arrow" has at least three parses, with "time," "flies," and "like" all potential candidates for the main verb. However, the interpretation where time is a substance which flies as an arrow does is so overwhelmingly more probable in normal speech that a probabilistic grammar should correctly identify it as the most probable, and therefore the most likely, and reject the others.

Probabilistic regular grammars are also known as Markov chains or Markov models. A *hidden Markov model* (HMM) is simply a Markov model where the grammar and probabilities are initially unknown and must be inferred from the observed sentences. Markov models have long been known to be inadequate for complete descriptions of natural language, as they fail to capture the necessary long-distance dependencies. On the other hand, they are simple to describe and easy to infer, and they do a very good job of describing local, short-term dependencies and patterns. Shannon's classic work on the entropy of English[42], for example, is based upon an implicit Markov model. They have found extensive use[13, 50] in speech recognition systems. More recently, Cutting et al.[15] used a HMM to tag English sentences with their parts of speech, while Stolcke and Omohundro[46] attempted to model language acquisition with a HMM as a first step to a general language acquisition problem[17]. And, of course, Brown[6, 7] used Markov models to try to develop automatic machine translation as described above in section 2.1.

*Probabilistic context-free grammars* (PCFGs) are of course more linguistically plausible, but also more difficult to work with. Here the sort of guidance suggested by the Marker Hypothesis has been used for a long time. Crespi-Reghizzi[14] and later researchers[2, 38] have studied so-called "structured samples" of languages and developed extensive algorithms to work properly with them. These samples consist of sentences in the language where some or all of the sub-constituents are marked by some form of explicit bracketing, such as "she ate (the hamburger) (with (a fork))." In the extreme case, this converts the process of grammar acquisition to the process of simply labeling unlabeled parse trees. Even with only partial bracketing, though, Pereira and Schabes[38] demonstrated vast improvement in the accuracy of the grammars inferred over those inferred from unbracketed corpora. The bracketing, then, which serves only to identify components, is an important and useful cue to grammatical identification. Marker words which also provide cues about which components they are marking should, then, be even more useful.

Psycholinguistic constraints from X-bar theory[26] have also been shown to help with the task of grammatical inference. Charniak[10] describes a system[9, 8] he built to identify PCFGs that focuses on the types of non-terminals that could appear as components within another non-terminal. For example, English pronouns can be difficult to tease apart from the English verbs by standard PCFG techniques. By imposing the single linguistic constraint that verbs were not allowed to appear in the expansion of a pronoun, the grammars inferred became much more accurate, understandable, and plausible. Being able to make this sort of modification clearly requires both a good understanding of English linguistics and of linguistic principles in general. Although not language-independent, it underscores the advantages to a system, for whatever NLP problem, which produces knowledge and rules encoded in a linguistically plausible and understandable format.

## 3 Formalizing the Process

The psycholinguistic properties described above should, in theory, be useful for any natural language problem. The current work focuses on their application to the translation of natural languages. Specifically, given a corpus of paired sentences, the Marker Hypothesis implies that there should be enough information in the corpus for a computer to extract the structural properties of both languages and be able to discern the necessary transformations to convert novel sentences from the source language into the target.

The mathematical basis for this program is described in section 3.1. It will be shown that any context-free grammar can be described in structures where all constituent structures (non-terminal symbols) are marked by the appearance of some terminal symbol. The Marker Hypothesis implies that such forms should not merely exist, but also be a natural and useful expression of human languages. By identifying these terminal symbols, a system could acquire a grammar for the desired language or languages and use it in further NLP work. Section 3.2 then describes the process of parameterizing the descriptions of the source and target language in such a way that they can both be useful for performing the translation task and can be identified automatically using standard multivariate optimization techniques. A complete description of these parameters, as inferred by the computer, constitutes the output of the system and can be used to translate novel sentences.

### 3.1 Marker-normal Form

As described in section 2.2, the crucial property for this work is the existence of identifiable classes of marker words.[4] Specifically, the formalism and system as described below assumes first that the languages of interest can be approximated by a context-free grammar, and second, that these languages can be naturally described by CFGs in *marker-normal form*, as defined below.

**Theorem 1** *To every CFG $\Gamma$ there corresponds an equivalent grammar in marker-normal form, where every production is of one of the following forms :*

$A \to \epsilon$

$A \to a$

---

[4]Or morphemes. The current work only focuses on marker *words*, but future developments will include morphological analysis from large corpora as a part of marker identification[27].

$$A \rightarrow A_0 a_1 A_1 a_2 A_2 \cdots$$
$$A \rightarrow a_1 A_1 a_2 A_2 \cdots$$

*(As usual, upper-case letters are nonterminal symbols, lower-case letters are terminal symbols, and $\epsilon$ is the null string of zero length.) All right-hand sides of productions are either a single terminal symbol, or are an alternating sequence of terminals and nonterminals.*

*Proof:* Greibach's theorem[24, 25] states that for every CFG, there exists an equivalent grammar in which all productions are of the form $A \rightarrow aBCDEF \cdots$ (so-called *Greibach-normal form*). For a grammar in this form, all right-hand sides consist of exactly one terminal symbol, followed by by zero or more nonterminal symbols. For any grammar of interest, begin by finding an equivalent Greibach-normal form grammar $\Gamma$ for it. This will then be transformed into an equivalent marker-normal form grammar.

Replace every production $A \rightarrow a\beta$, where $\beta$ is a string of two or more nonterminal symbols, with two productions involving a new nonterminal : $A \rightarrow aX$ and $X \rightarrow \beta$. At this point, all productions involving the original nonterminals of $\Gamma$ are in the required form for marker-normal form.

Now, consider a variable $B$ that appears in the right-hand side of a rule $X \rightarrow \beta$. If $B$ is the left-hand side of several production rules, create multiple production rules for the nonterminal $X$ with the right-hand side of each production rule for $B$. Repeat this process with the Cartesian product of all original nonterminal symbols of $\Gamma$. At the end of this process, every rule of the original grammar with multiple nonterminals has been replaced with a rule of the form $A \rightarrow aX$, with a single (marked) nonterminal variable. As the right-hand side of $X$ did not contain any of the new nonterminals, every nonterminal in has been replaced by a marked nonterminal, and so the right side of every $X$-production is also completely marked.

To convert this entirely to marker-normal form may require the addition of another nonterminal between two terminals in the right-hand side of a production. Simply add the rule $\Omega \rightarrow \epsilon$, for a novel nonterminal $\Omega$, and replace all such right-hand sides $\gamma$ with $\Omega\gamma$ creating the initial nonterminal as required. This grammar is clearly in marker-normal form and also clearly equivalent to the Greibach-normal form grammar from which it was derived.

An example of this transformation may be useful. Consider this grammar, which generates sentences that consist of one or more sequences of balanced parentheses around a '+' character.

$$S \rightarrow (E)S \quad S \rightarrow (E)$$
$$E \rightarrow (E) \quad E \rightarrow +$$

Converting this grammar to Greibach-normal form produces :

$$S \rightarrow (EPS \quad S \rightarrow (EP$$
$$E \rightarrow (EP \quad E \rightarrow +$$
$$P \rightarrow )$$

All productions are already in marker-normal form except for the rules $S \rightarrow (EPS$, $S \rightarrow (EP$, and $E \rightarrow (EP$. Create two new nonterminals $X$ and $Y$ such that, respectively, $X \rightarrow EPS$ and $Y \rightarrow EP$.

Making this replacement, the (new) grammar becomes :

$$S \rightarrow (X \quad S \rightarrow (Y$$
$$E \rightarrow (Y \quad E \rightarrow +$$
$$P \rightarrow ) \quad X \rightarrow EPS \quad Y \rightarrow EP$$

As $S$ and $E$ both have two possible productions, $X$ has four first-level expansions and $Y$ two. Performing these yields

$$S \to (X \quad S \to (Y \quad E \to (Y \quad E \to +$$
$$P \to )$$
$$X \to (Y)(X \quad X \to (Y)(Y \quad X \to +)(X \quad X \to +)(Y$$
$$Y \to +) \quad Y \to (Y)$$

and a similar substitution should be performed for $Y$. Finally, adding the rule $\Omega \to \epsilon$ and padding with $\Omega$ as necessary yields a final version in marker-normal form :

$$S \to (X \quad S \to (Y \quad E \to (Y \quad E \to +$$
$$P \to ) \quad \Omega \to \epsilon$$
$$X \to (Y)\Omega(X \quad X \to (Y)\Omega(Y \quad X \to +)\Omega(X \quad X \to +)\Omega(Y$$
$$Y \to +\Omega) \quad Y \to (Y)$$

The reader may already have noticed that the original grammar for this example was already in marker-normal form. Although accidental, this is not coincidental. Most of the languages and grammars small enough to present as an example have an easy, quick, and understandable presentation in marker-normal form, and it can be a difficult task to find or develop a grammar that does *not* have such a presentation. Although the construction did produce a grammar in marker-normal form, it was considerably more complex, largely due to the added complexity from putting it into Greibach- normal form. This suggests two things : first, that a more elegant proof to theorem 1 exists, and second, that people tend to think in terms of marker-normal-form languages—and therefore this presents more evidence for the psychological utility of the Marker Hypothesis.

Theorem 1 has an immediate corollary to reduce the necessary size of the production rules, at the expense of the number of such rules :

**Corollary 1** *To every CFG there corresponds an equivalent grammar form, where every production is of one of the following forms :*

$A \to \epsilon$

$A \to a$

$A \to A_0 a_1 A_1$

$A \to A_0 a_1 A_1 a_2 A_2$

*Proof:* Exercise 4.11 of [24, p. 66] states that every context-free language can be generated by a grammar of the form

$A \to a$

$A \to aB$

$A \to aBC$

The construction of Theorem 1, when applied to the above grammar, produces a grammar of the desired form. If necessary, an $\epsilon$-generating non-terminal symbol can be prepended to any production rules that do not already begin with one.

This method clearly results in much larger and potentially less-coherent grammars than the more standard Chomsky- and Greibach- normal forms. However, the Marker Hypothesis implies that explicitly marked grammars such as these are more psychologically plausible and thus that these grammars are likely to be more natural and understandable for human languages. In particular, natural language should tend to have relatively simple descriptions in which the set of terminal symbols that appear alone in productions is distinct from the set of terminal symbols that appear in a marking context; in other words, that the set of marker words is distinct and identifiable.

### 3.2 Inference by Grammatical Optimization

Within this framework, there is at least one straightforward approach for language/grammar acquisition. Given a set of training sentences one can automatically[45] extract a set of candidates to be "marker words". The system can then examine all sentences to determine the particular marker words which separate the components (non-terminal variables) of the full sentences as well as to determine how many and which non-terminals are relevant to the production of full sentences. A similar sort of analysis can be performed for the various sub-components. In principle, this sort of analysis can be successively refined until it describes the training set and by extension the source language, to any desired degree of accuracy.

To complete the task of self-organizing automatic translation, it is then necessary to convert the training sentences as well as novel source sentences into their translations with the same high degree of accuracy. The main problems here are twofold. First, the lexical items used to create the source and target language are different; highly correlated, but not simply a one-to-one mapping. Second, the structure of the source and the target languages may be different. For example, the source language may be English, where the majority of sentences follow the pattern subject-verb-object. The target language may, in turn, have the basic pattern verb-subject-object. To convert sentences from English to the target language requires a systematic restructuring of the components, placing the verb first and the subject second.

The first problem can be solved by the development of a context-sensitive translation dictionary. A certain number of potential lexemes and compound words will be designated as "lexicalized", by which read, they are not further broken down by the analysis process and instead translated as a unit. In most cases, these lexicalized subunits will be single words that will be translated into other single words. However, the deletion of words (such as articles, when translated into languages that do not have such distinctions) can be accomplished by "translating" the undesired word into $\epsilon$, and words can be added by translation of $\epsilon$ into something more useful. Context sensitivity can be added by the development of multiple translation dictionaries, each associated with a particular syntactic structure, that define a word's translation to be different in different structures. These dictionaries can be compiled automatically, without human intervention.

The second problem, that of restructuring the translation to reflect the structure of the target language, can be accomplished by a permutation of the relevant components of the source sentences. These components are of course identified in the analysis of source sentences. These components can then be assembled (concatenated) in a different order. This order, although in theory independent of the source language, can instead be described as a permutation of
the source ordering for a given source→target pairing. These permutations, of course, can be
learned individually, or can be identified in larger units reflecting various universals of word
order.

One can see from the above discussion how an appropriately chosen set of parameters can
accomplish the translation task. Specifically, for the system described here, these parameters
are :

- a marker-normal form CFG for the source language,
- a set of context-sensitive dictionaries,
- a set of permutation relations for every rule in the CFG.

The following section describes a simple prototype system for finding such parameters from
untagged, unanalyzed bilingual corpora using standard multivariate optimization techniques.

## 4 METLA-1, a computational prototype

The formalism described above has been tested on a small-scale with a program called METLA
(Machine Engineered Translation by Language Acquisition) on a number of small data sets
involving various languages. METLA-1 itself is a relatively small C program with a number of
simplifying assumptions built in. In its general operation, the system accepts a set of sentences
with the "appropriate" translations and builds from that an analytic grammar of the source
language with appropriate translation information to produce the example target sentences from
the example source sentences. The only other information provided is a list of vocabulary items
in both languages, in no particular order. This simplifies the programming task by allowing the
program to examine numbers (word #14 translates to word #61) rather than explicit strings
with the attendant storage and processing costs. This is a convenience which can be easily
added by preprocessing but does not actually add anything to the power of the system.

The system itself works by assuming a basic, randomly parameterized transfer function.
Initially, the system generates a random context-free grammar with random marker words, a
set of random dictionaries, and a random permutation set for each CFG production. It then
uses a standard optimization technique to tune the parameter set, in this fashion determining
the set of parameters which produce the best translation—"best," in this case, being defined
as the translation that produces the closest match between the target (example) sentences and
the translated source sentences.

The grammar itself is characterized as a fixed set of $N$ rules, numbered from zero to $N-1$.
Each of these rules has a fixed fanout of non-terminal symbols $k$, so every rule in the grammar
is of the form

$$A_i \rightarrow A_x m_{i,1} A_y \cdots m_{i,k-1} A_z$$

where each $A_?$ is a non-terminal in the set $A_0 \ldots A_{N-1}$ and each $m_{?,?}$ is a marker word (selected
from some context-dependent set of marker words) that marks the separation between the
various constituents of $A_i$. $A_0$ is designated as the starting symbol of the grammar, and the
example sentences are parsed in a strict top-down fashion, partitioning each sentence into its
constituents at the appearance of the leftmost element of each marker set, in order of appearance in the rule of grammar. These constituents are then recursively parsed in accordance with the single rule corresponding to their nonterminal, and so on until the sentence has been broken down into only lexicalized items. These items are, of course, translated using the various context-sensitive dictionaries the system has developed. Figure 1 shows an example of this parsing scheme in action, using an actual grammar developed by the system in the course of the English→Urdu experiments. Each stage of the parsing is described twice, once in a graphical tree format and again in a more compact text-based format where constituent boundaries are marked by parentheses.

At this point, the translated utterances must be reassembled in keeping with the structure of the target language. Associated with each analysis rule is a permutation of its components, which describes the order in which the translated components are to be concatenated to form the larger translation. This permutation-concatenation process is carried out up the recursion, until at the end, rule $A_0$ terminates with a complete translation of the source sentence, which is measured against the listed target sentence for similarity. Depending upon the similarity, any or all of the translation scheme may be modified slightly until the desired overall accuracy has been reached. Figure 2 illustrates this process in detail on the sample sentence from figure 1.

Because of the strict top-down nature of this parsing as well as the fact that each nonterminal is associated with only one production rule, the nature and accuracy of the grammars that METLA-1 can infer is strictly limited. In addition, because of the focus of METLA-1 only on marker *words*, many morphologically marked structures will not be found. Finally, the system starts from a randomly-chosen starting grammar and vocabulary, rather than using any techniques to select a good starting point, which slows the inference task considerably.

The inference itself is done by a standard multivariate optimization algorithm called simulated annealing[33, 31]. This technique was originally designed as a model of crystal growth and metal annealing, but has found widespread use in a variety of contexts and has the advantage of being well-known, well-studied, and reliable. In simplest terms, it is a variant of a random walk through the event-space of interest. At each step, the algorithm considers a random change to the set of parameters (for example, changing a single nonterminal symbol in a particular rule, adding or deleting one word from a particular set of marker words, or changing the translation of one word in a particular dictionary) and measures the quality of the translations produced by that set. If the changed parameters result in improved performance, the system accepts the new parameter set for further work. Even if the parameters reduce performance a bit, the system *may* still accept the new parameters as long as the performance loss isn't too great. As the algorithm progresses, the notion of "too great" is gradually tightened until the algorithm accepts only improving moves and eventually will find the global performance maximum.

Automatic measurement of translation performance can be difficult to perform. Many psycholinguistically plausible measurements are computationally expensive or technologically impossible. However, the sort of tasks that are computationally viable may produce "false positives" which appear to be related to the correct translation but in fact are very different. (Consider the effect of adding or deleting a 'not' from an English sentence.) After several experiments, the system uses a modified greatest-common-subsequence formalism[36], which should be familiar to most UNIX programmers as the *diff(1)* algorithm. Specifically, this measures the number of changes (insertions or deletions) that distinguish the translated source sentence from the desired target sentence. Sentences are thus graded on the number of words that would need to be added to or deleted from them to produce the exact form in the examples, an approximate measurement of the amount of work a human editor would need to do.

"the man is in the shop"               the man is in the shop

Initial rule divides at the first appearance of a determiner and leaves this sentence unchanged.

"the man is in the shop"               the man is in the shop

Second rule divides immediately before the appearance of the copula ('is', etc.)

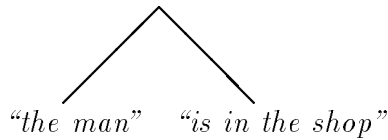
(the man) (is in the shop)

"the man"    "is in the shop"

The second part of the above division is now divided before the appearance of a preposition

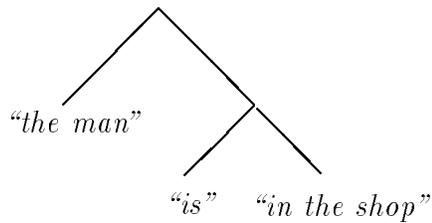
(the man) ( (is) (in the shop) )

"the man"
"is"    "in the shop"

And so forth (dividing immediately before the existence of a determiner)

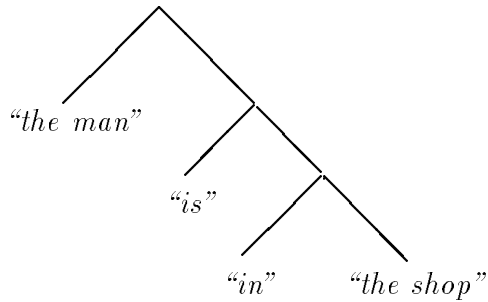
(the man) ( (is) ( (in) (the shop) ) )

"the man"
"is"
"in"    "the shop"

**Figure 1**: Example English parsing (derived from E→U)

"the man is in the shop"

This can be partially parsed into

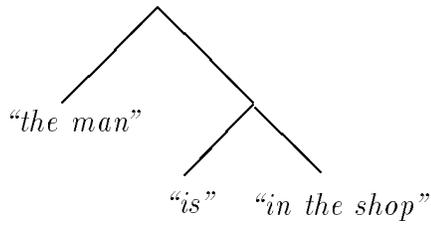

(the man) ( (is) (in the shop) )

Each of these phrases is individually translated (into Urdu)

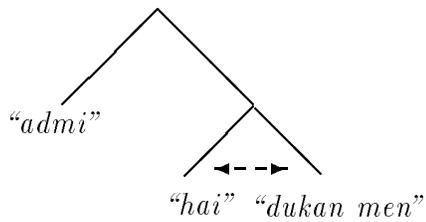

(admi) ( (hai) (dukan men) )

And the children at each node may be permuted to form the final structure

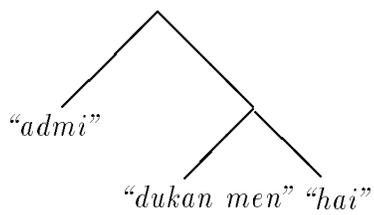

(admi) ( (dukan men) (hai) )

which is concatenated to form the final output string

"admi dukan men hai"     admi dukan men hai

**Figure 2**: Example of METLA-1 translation algorithm

To recap, then, the METLA-1 translation system is a small-scale inference system using the linguistic formalism described in section 3.1 as applied to machine translation. Starting from a random skeletal grammar, the system uses a simulated annealing to simultaneously set a large number of parameters to define the parsing, lexical transformations, and structural changes necessary to transform sentences in the source language into the target language. The inputs to the system are a set of translation examples, and the output is a parameter list describing the inferred transfer function. This set of parameters constitutes a parsing and translation system that can be used in a separate, standalone machine translation system. Internal measurement of the parameter sets (within the annealing) can be done in a number of ways, but the current system uses a variant of the the UNIX diff program to measure the approximate amount of work a human editor would need to do to clean up the machine-translated outputs. The system should work in a variety of linguistic and functional contexts without any major modifications. For example, to change the system from an English→French translator to an English→Japanese translator should require only a new set of training data and no changes to the source code.

## 5  Experiments

### 5.1  Experimental Descriptions

METLA-1 has been implemented and tested on a variety of small problems. This report covers the most interesting from a psycholinguistic point of view. The two problems posed to the system were English→French and English→Urdu translation from small unanalyzed corpora. The system was allowed to anneal on the input corpus for a fixed length of time which varied from trial to trial, and performance was then measured (by hand) on both the training data and a set of independent test data. In addition, the transfer functions that the system had developed were examined carefully to determine exactly what words, classes, and properties the METLA-1 system found useful for translation.

In the first experiment, the system was asked to infer the transfer function to perform English→French translation from a small artificial corpus (c. 30 full sentences) developed by native speakers of English and French working together. Although this grammar was very simplified in many regards (for example, all sentences were in present tense), there were nevertheless a number of systematic differences that were exploited. For example, French nouns have gender, while English nouns typically do not, and it was expected that this would be a difficult distinction for the system to draw. Similarly, words like 'that' have several different translations into French ('ce,' 'cette,' 'que', ...), depending upon the grammatical context in which they occur. The test data were produced by handing an English vocabulary list to a (different) native speaker of English, with a brief description of some of the major limitations of the training grammar ("All sentences are in present tense, and all nouns are singular.") and asking for example sentences in English. Of course, in some cases this resulted in word usage that was not covered by the training grammar, but this is a risk in any example-based translation system. The test data used were 47 sentences of which ten had not been covered by the input grammar. The system was run several times on the training data with different random starting points, different allowed annealing times, and different (small) grammar sizes, and each time the errors made were comparable and the results of about the same quality.

The second series of experiments was an attempt to compare the performance of the system against a related task for humans, that of learning a foreign language from instructional text.

In addition, it measured the approximate language-independence of the designed system. The system itself remained the same in both sets of experiments, although some parameters (length of annealing, maximum number of non-terminal symbols per production rule[5], etc.) were changed in the various experiments. The only necessary change, then, to convert METLA from a translator to French to a translator to Urdu was to change the input corpus. For these experiments, the system was presented with the complete text of the example sentences and vocabulary list from lesson 2 from [49]. The language Urdu was chosen for two reasons. First, it is a language with several major structural differences from English, being of basic word order subject-object-verb instead of subject-verb-object, having postpositions instead of prepositions, and having no notion of the definite/indefinite article distinction[6]. Second, both instructional texts and native speakers were available for cross-validation. The text of the lesson covered imperative sentences ("Put the letter on the table.") and copula-locatives ("The hat is in the office."), with a complete vocabulary list and seven examples of full sentences. As above, the experiment was repeated several times and all results were substantially similar.

## 5.2 Evaluation

One of the difficulties involved with the development of a machine translation system is the evaluation of the end product. Is it better, for instance, to produce an ungrammatical translation that nonetheless seems to capture the meaning of what the original said, or to produce a grammatically flawless sentence that states something completely different from the original? How should the system respond to unusual, metaphoric, or ungrammatical inputs?

### 5.2.1 Black Box Evaluation

For many fully self-automated translation systems (e.g. [6]), the problem can be made worse by the relative opacity of the inferred translation system. There is no easy way to examine the internal workings of the algorithm to determine the nature and causes of a translation error or to identify how to repair the error. And for translation systems using Markov models and similar oversimplified grammatical structures, it may not be possible to understand the cause of the error even after a lengthy and extensive analysis of the translation parameters, as the underlying model is too distant from people's intuitive understanding of how languages are put together.

Nonetheless, it is possible to do some sort of a black box analysis of the output of the system. Brown et al.[6], for instance, performed their analysis on the basis of hand-classification of sentences into five types, ranging from "Exact" (Identical to what the Hansard translator chose), through "Alternate" (Different phrasing but the same idea expressed), down to "Ungrammatical." This sort of hand-classification for final system evaluation is useful because it directly measures the appropriateness of the final product in a way that more automatic measures (such as *diff*) cannot. For the METLA-1 prototype, though, this particular classification was less useful than the classification actually used. Because of the limited vocabulary and grammar in the experiments, very few different grammatical ways to express the same idea were available. It was therefore more useful and appropriate to classify sentences (again by hand) into the categories "Correct," "Minor errors," and "Gibberish." The first category corresponds to "Exact," above. The third category describes sentences that were so syntactically ill-formed as to be unintelligible and would be a subset of Brown's "Ungrammatical" sentences. The second category would be classified by [6] sometimes as "Alternate" and sometimes as "Ungrammatical." These tend to be syntactically invalid but semantically understandable. They also tend to reflect (subtle) properties of the *source* language that are slightly changed in the target language. In fact, they closely resemble typical errors of first-year language students. Examples of these from the English→French experiments include deletion of sentence complementizers[7], deletion of reflexive particles, or gender errors.

---

[5]i.e. the maximum fan-out
[6]i.e. there are no words corresponding to 'the' vs. 'a.'

When this sort of analysis is performed on the results of the English→Urdu experiments, the system learned the original training corpus (the example sentences from the lessons) perfectly and could reproduce it without errors. Testing on novel sentences (the exercises) revealed 72% completely correct, and only 7% translated as "gibberish." Upon further analysis (see section 5.2.2), the training corpus was shown to be unrepresentative of the test corpus, and in particular was missing coverage in context for several words. When the training corpus was updated to include coverage for the missing items, the system could still learn the training corpus perfectly and the percentage correct on novel items of the same forms increased to 100%.

The English→French experiment, because of the higher syntactic complexity in conjunction with the limited scale of the prototype, performed less well overall. Typical performance for the system on the training corpus was approximately 61% correct, 29% minor errors, and only 10% gibberish. On the test data, performance was lower, with only 36% correct, 21% minor errors, and a full 44% gibberish. However, when the test sentences that presented structures unrepresented in the grammar were excluded, the performance improved, up to 41% correct, 19% minor, and 41% gibberish. These numbers can be compared with the results from [6], where an early version of the system was able to correctly translate 48% of the test data based on a much larger training (and testing) corpus.

Clearly, much additional work will be required before METLA turns into a commercial-quality translator. However, given the known structural limitations of the implementation and the small grammars that it used for these experiments, these still represent a significant accomplishment in the development of a psycholinguistically plausible MT system. Perhaps equally significantly, to convert the system from one language to another required approximately an hour of human effort to type in the training data, and no system modifications. This indicates that language-independent induction of transfer functions may be a viable approach to machine translation.

### 5.2.2 White Box Evaluation

A major advantage of a psycholinguistically plausible approach is that, if properly done, the output of the system can be directly converted into a grammar and dictionaries for the appropriate languages. This makes it possible to directly analyze the plausibility and appropriateness of the various transfer rules and to improve them by human intervention. Some of the simplifications made in the course of developing METLA-1 have made it more difficult to perform this task, but one can still examine the source grammar and transfer functions which the system developed and use this information to change the transfer rules or training data.

---

[7]The 'that' in the English sentence "I believe (that) rocks sink" is optional. The corresponding 'que' in its French translation is required.

For example, in the English→Urdu experiment, the training data consisted of copula-locatives ("the hat is on the chair", "the man is in the shop") and imperative sentences ("wait in the office," "send the knife to the house"). Each of these had to be rearranged into verb-final form, and the prepositions had to be converted to postpositions. In addition, all the determiners ('a,' 'the,' 'this,' etc.) needed to be deleted, so the final result of translation would be something like the word-for-word translation of the string *"knife house to send."

Upon examination, the word classification and translation methods make sense. For an example, one of the early experiments initially divided all sentences into two parts based on the first appearance of a determiner or preposition. This divided imperatives ("wait in the office") into their verb components followed by one or more arguments which were translated by another set of rules. The translation of the verb was permuted to follow the rest of the sentence, giving the necessary verb-final form. On the other hand, declarative sentences ("the book is on the table") are passed through this initial rule unchanged, to be divided later at 'is' into subject, verb, and location, and permuted appropriately. This sort of analysis can be carried out to any desired level of detail.

Even this simplified analysis, however, is enough to demonstrate the advantage of a psycholinguistically plausible and symbolic representation. The statement "to be divided later at 'is'" is, in point of fact, slightly inaccurate. Using the first version of the training data, the system accurately inferred that 'is' serves to mark the boundary between subject and verb. However, it also inferred (wrongly) that 'knife' and 'man' were also part of that same marker group. This resulted in a small number of incorrect translations of the testing sentences.

Further examination of the input corpus showed the reason that these errors had been made. Although the system was presented with a full vocabulary list ('man'/'admi', 'house'/'ghar', and so forth) of individual words, only a subset of those words had been presented in the context of a phrase or sentence. Although the system, then, had learned that 'man' translated to 'admi,' it had no evidence about the part of speech of 'man.' The system had no way of knowing, for example, that the word 'man' was not an alternate form of the copula. In general, the lists of marker words are obviously of one or more grammatical classes, with potentially a few outliers that represent words that have never been seen in that context and therefore may or may not be relevant. With this observation, it becomes/became obvious that the input examples were not representative of the testing data, and that some new input was required. After adding two more sentences to provide context for these words, the percentage correct increased in later experiments to 100%.

Similar analysis can be done for the more grammatically-complex English→French experiments. Because of the greater syntactic complexity, the system as built proved to be oversimplified in several important regards and some errors were in that sense inevitable. For example, consider the following set of sentences :

(the man) (kisses) (the woman)
(le homme[8]) (embrasse) (la femme)

(the man) (kisses) (her)
*(le homme) (embrasse) (la)

---

[8]The process that converts, for example, 'le homme' into 'l'homme' is almost purely phonological and was ignored for simplicity in the METLA-1 system.

| |
|:---:|
| bring the letter from the shop |
| (bring) ((the letter) (from the shop)) |
| (lao) ((chitthi) (dukan se)) |
| chitthi dukan se lao |
| wait in the office |
| (wait) (in the office) |
| (thairo) (daftar men) |
| daftar men thairo |
| put the box on the table |
| (put) ((the box) (on the table)) |
| (rakho) ((sanduq) (mez par)) |
| sanduq mez par rakho |

Table 1: Sample English→Urdu translations with partial analysis

*(the man) (her) (kisses)
(le homme) (la) (embrasse)

In the first pair of sentences, the pattern subject–verb–object is used in both French and English, so the identity permutation is appropriate. In the second and third pairs, the pattern becomes subject–object–verb in French, so the identity permutation is no longer appropriate. However, as each non-terminal symbol (sentence, in this case) has only one rule and one permutation associated with it, the system is forced to select one and only one of object-final or verb-final structure.

On the other hand, the system correctly learned appropriate translation structure for a large part of the input corpus. For example, the original sentences are parsed into three pieces based upon the existence first of a verb, and then of a determiner or pronoun. Noun phrases (which begin with a determiner in the input corpus) are themselves partitioned into classes of masculine/feminine noun phrases so that the gender of the determiner is correctly set.

The major error made by the English→French system was that it found a local maximum in reusing one of the production rules. Because any translation system should allow for recursive structures ("John said that Mary told him that Susan said that ..."), the system is permitted to call rules that have already been called. The system tended to find a local maximum where the rule used to separate masculine from feminine nouns was the same rule used to parse the original sentence, and so it conflated the two categories of verbs and feminine nouns. This meant, in turn, that sentences such ase "that woman washes a car" were divided not as "(that woman) (washes) (a car)" but instead as *"(that) (woman washes) (a car)." This error could presumably be rectified by allowing the system to use more production rules, but is more appropriately solved by a better parsing algorithm in general.

Some sample results are attached as tables 1 and 2. Each table shows a number of sample sentences (in the nearly opaque parenthesized format) along with their primary division into constituents. the translations of those constituents, and the final translation after it has been permuted and concatenated.

The errors in table 2 should be explained. First, note that the division of the third sentence is incorrect—"the man that touches the car" is an entire component and the main verb of the sentence is the *second* token of 'touches.' This is an artifact of the admittedly broken METLA-1 parsing algorithm, which divides at the first appearance of a given token. That this sentence is correctly translated at all is a tribute to the remarkable structural similarity between this sentence and its French translation. The fifth sentence is an example of a so-called "reflexive" verb; the proper translation should be "ce chat se lave," where 'se' is a general pronoun meaning 'self.' In English, certain verbs can be intransitive when the subject and object of the verb are the same—for example, "I shave (myself) every morning," "I wash[9]," and so forth. Some of these verbs, in turn, *must* be expressed with the reflexive particle in French but with an ordinary direct object otherwise. This leads, in turn, to another example of the multiple-necessary-permutation problem discussed above.

| |
|---|
| the glass touches a car |
| (the glass) (touches) (a car) |
| (le verre) (touche) (une voiture) |
| le verre touche une voiture |
| she washes a cat |
| (she) (washes) (a cat) |
| (elle) (lave) (un chat) |
| elle lave un chat |
| the man that touches a car touches a glass |
| (the man that) (touches) (a car touches a glass) |
| (le homme qui) (touche) (une voiture touche un verre) |
| le homme qui touche une voiture touche un verre |
| that man washes a car that she creates |
| (that man) (washes) (a car that she creates) |
| (ce homme) (lave) *(une voiture qui elle creee) |
| *ce homme lave une voiture qui elle creee |
| this cat washes |
| (this cat) (washes) () |
| (ce chat) (lave) () |
| *ce chat lave |

**Table 2**: Sample English→French translations with partial analysis

The fourth sentence is more interesting. The word 'qui' in the fourth example sentence is a relative pronoun used only for people (like 'who'). As an inanimate object, "a car" should have taken the relative pronoun 'que' as a translation of 'that'. However, notice should be taken of the mistake that the system did not make. The other token of 'that' in the sentence was a demonstrative determiner, which was correctly translated as 'ce', taking into account the gender of 'man'. The system correctly identified the second 'that' as a relative pronoun and not a demonstrative determiner. Similarly, the third sentence indicates an ability to distinguish between feminine nouns ("une voiture") and masculine ones ("un verre"), a relatively subtle grammatical point. These results, then, indicate an ability on the part of METLA-1 to determine remarkably small grammatical structures and to appropriately account for and to produce them as needed in the translation process.

---

[9]In some dialects, not including the author's, this concept would be expressed as "I wash up."

# 6  Needed Improvements

As has been stressed repeatedly above, METLA-1 is an exercise in proof-of-concept. Its current performance is hardly surprising in light of the number of simplifying assumptions that were made in the course of development. It is therefore reasonable to try to determine what sorts of improvements will be necessary in future versions of the system. In general, the areas that need improvement or at least further study can be summarized as follows :

- improvement of parsing algorithm

- replacement of simulated annealing as the optimization/learning algorithm

- incorporation of morphological analysis

- incorporation of additional psycholinguistic principles and constraints

- automatic alignment of source corpora

- preprocessing for better starting point for the learning algorithm

- incorporation of part-of-speech tagging

- scalability of this general approach

The first and foremost candidate for improvement is the parsing algorithm. As has been pointed out in section 5.2.2, the limitation of a single rule per nonterminal symbol and the absence of any sort of bottom-up, data driven parsing scheme forces errors to occur in syntactically complex data. Some sort of data driven parser, such as the chart parser of [28], is obviously necessary.

Using simulated annealing for learning has its limitations. First, it tends to be slow. Second, it does not lend itself well to parallelization. Third, it is difficult or impossible to extract meaningful partial results at the half-way point of a long annealing session. Fourth, and perhaps most importantly, it is not incremental. An ideal system would be able to build a solution on the data available today, and then modify that original solution by comparison with the data available next week without losing what was originally learned. This would lead, for example, to a system that could learn to translate by optimizing, chapter by chapter, over a standard second-language textbook. Unfortunately, this sort of incremental learning is beyond the capabilities of a system based on simulated annealing, and so a new and improved optimization algorithm would be useful.

There are other multivariate optimization techniques that could be plugged in, such as tabu search[19, 20] and genetic algorithms[4]. Some preliminary work has been done on incorporating these techniques, but the results have been merely comparable to the work presented here. Whether this results from poor implementation of the genetic algorithms or is simply a statement about the appropriateness of general-purpose optimization techniques remains unresolved. Even with the mediocre results, however, there are clearly some limitations of simulated annealing.

For the small problems that have been studied up to this point, it is not really a problem that there is no morphological analysis. If there are only 5 (or 30) verbs in the system vocabulary, the system can reasonably be expected to learn and classify them by enumeration. As the vocabulary becomes larger and larger, though, the number of words in the open-class categories such as verbs, nouns, and adjectives, will grow beyond the capability of any system to learn them all by enumeration. However, a system with the ability to identify and classify words by morphological analysis, or simply to segment novel words into morphologically plausible sections, would be able to correctly parse and translate a much larger set of sentences. For many languages, there already exist morphological analyzers. For the languages that don't, induction of morphological regularities has been studied by several people; Borin[5] describes a system based upon several linguistic theories; I propose elsewhere[27] a statistical approach based upon Shannon's[41, 42] notion of information theory. Any of these tools could be incorporated as a preprocessor into future versions of the system.

Many available "linguistic universals" have not been incorporated into the METLA-1 system. The Greenbergian universals about word order, for example, were alluded to briefly in section 2.2, but have not been directly incorporated into the design of the system. Even as simple an observation that "all languages have verbs and nouns" could potentially be useful for system improvements. Greenberg[23] proposed relationships between how a language structure things around nouns and how the same language structures things around verbs—by limiting the grammars to those compatible with Greenberg's universals, the inference task could be speeded up. Similarly, the X-bar convention, as proposed by [11, 26], suggests that the head/dependent distinction (as between nouns and full noun phrases) may be a linguistically important notion, and that this should be incorporated into the grammar.

The reliance of the system on paired sentences, so-called "aligned data," may also be problematic. Although large bilingual corpora exist, the sentences in these corpora do not tend to equate in a nice, linear, one-to-one fashion. Any practical use for this system, then, would need as a preprocessor a program or module that can take a bilingual corpus and produce an aligned bilingual corpus. Fortunately, such systems are being designed and built; see [18, 29, 51] for examples. Similarly, it would be useful to have a preprocessor build a rough guess at a dictionary before the system itself actually started working on it (especially with some sort of incremental learning technique)—by simply calculating the correlation among the various words in the input and output corpora, a dictionary with a relatively high accuracy can be built and used as a starting point for further refinement.

Part-of-speech tagging, as typified by [15], is another well-studied area in statistical linguistics. It is not clear whether or not the system would be improved if the source and/or target corpora were tagged by such an automatic system. Similarly, the PCFG framework as typified by [38] is a well-known paradigm. It may or may not be useful to incorporate some notion of probability into the parsing or translation algorithms to improve the performance in the presence of ambiguous data. Finally, any AI project must be evaluated carefully to determine how well it will scale to real-world applications with huge amounts of relevant data. Any improvements to the system to improve its scalability would be useful.

Finally, it is, or should be, clear that any EBMT system that relies upon a translation database of only fifty sentences is inadequate. Not only is efficiency of scaling a concern, but the question of whether or not this approach scales at all should be addressed. Much further work is needed, either with METLA-1 or a future system, to test it in a real-world context with real, large-scale corpora. Without this sort of testing and development, METLA-1 is only a toy system, or a prototype.

# 7  Conclusions

To solve linguistic problems, one needs to understand linguistics. Twenty years ago, that statement would have been uncontroversial. Today, with faster computers, larger disks, and greater memory available, scientists can work more directly with examples from linguistic data. This has led some researchers to use methods that focus more on computational efficiency (e.g. hidden Markov models) and can sometimes be linguistically naive or use very language-specific structures. The work presented above is an attempt to graft psycholinguistic principles onto the EBMT framework in a language-independent fashion. Rather than using human effort to develop an exhaustive analysis of the source and target language, the human effort can be put into identification and incorporation of linguistic principles such as the Marker Hypothesis[22], X-bar theory[26], and structural universals about language[23]. By focusing on the same sets of principles that linguists use to describe novel languages, the same system could be used for many different language pairs, addressing exactly those inter-language differences that linguistic typologists find interesting.

METLA-1 is a prototype system developed to address and test some of the concepts necessary to produce such a language-independent analysis system. Focusing on a restricted set of linguistic universals (primarily the Marker Hypothesis) and on small sets of data, it nonetheless manages to produce respectable performance on the structural analysis, transformation, and translation of novel sentences. In addition, the structures and grammatical classes used are logical and linguistically sensible—for example, the system picks up readily on the concept of prepositions, correctly gathers the prepositions together in a class, and identifies that the dependent noun of a preposition follows the preposition itself in English.

This work does not exclusively focus on grammatical induction. Although grammatical induction is an important part of the task, neither the problem (translation) nor the approach guarantees that the system will learn anything usable for grammaticality judgements. For a simple example, a system trained to translate (US) telephone numbers from English to French would not necessarily learn that telephone numbers are comprised of seven digits, divided into groups of three and four by a hyphen. At the same time, the system would presumably be robust enough to translate malformed phone numbers without causing system errors. This is clearly an advantage in dealing with real-world input, where typographical errors and misphrasings are not uncommon. At the same time, this system will include grammatical structure which should result in more robust, understandable, and linguistically plausible translation functions than the Markov chains developed by [6].

Finally, although this system uses examples to develop its translation functions, there are several crucial differences between METLA-1 and the more mainstream EBMT paradigm. First, other than the notion of paired sentences, there is no preanalysis of the translation database, which greatly reduces the load on the human developers of the system. This system also produces a reduced database, explicitly extracting patterns from the example database rather than finding them as needed in on-line examples.

The results of the tentative experiments indicate that induction of transfer functions from untagged, unanalyzed bilingual corpora is a computationally and linguistically viable task. Furthermore, the addition of linguistic information into the algorithm itself produces more understandable and thus maintainable results. In particular, these results seem to show that hours, days, or months of computer time can be substituted for the time of human translators if the appropriate low-level bilingual corpus is available. The METLA-2 system and its descendants, with some of the major weaknesses fixed, will hopefully be able to achieve a high degree of performance on any and all language pairs, with a great linguistic fluency and maintainability.

## 8 Acknowledgements

The first person to be acknowledged, of course, would be my advisor, Jim Martin, as no graduate student can accomplish *anything* without such advice and support. I would also like to acknowledge the members of my committee, in particular Lise Menn, for her valuable discussions about psycholinguistics; Wayne Citrin, for his engineering help; Michael Main, for assistance in formalizing what eventually became Theorem 1; as well as an unrelated faculty member, Karl Winklmann, for similar help with the formalism. This work would not have been possible without the help of my various native speakers, and I am grateful to Nathalie Bonnardel, Bhavna Chhabra, Anand Dhingra, and Kumiyo Nakakoji for their respective expertise. I would like to thank Jerry Feldman of ICSI for the opportunity to present a preliminary version of these results to his team, and Andreas Stolcke, Jonathan Segal, and Dan Jurafsky for their feedback from that presentation. Finally, I would like to thank Sonia Connolly, Dan Carroll, and Alex Popiel for their engineering assistance and valuable discussions.

## References

[1] Dana Angluin. Inductive inference of formal languages from positive data. *Information and Control*, 45:117–35, 1980.

[2] J. K. Baker. Trainable grammars for speech recognition. In Jared J. Wolf and Dennis K. Klatt, editors, *Speech Communication Papers for the 97th Meeting of the Acoustical Society of America*, New York, 1979. Algorithmics.

[3] Brent Berlin and Paul Kay. *Basic Color Terms : Their Universality and Evolution*. University of California Press, Berkeley, CA, 1969.

[4] Albert Donally Bethke. *Genetic Algorithms as Function Optimizers*. PhD thesis, University of Michigan, January 1981.

[5] Lars Borin. The automatic induction of morphological representation. Reports from Uppsala University, Linguistics (RUUL) 22, Department of Linguistics, Uppsala University, Uppsala, Sweden, 1991.

[6] Peter F. Brown, John Cocke, Stephen A. Della Pietra, Vincent J. Della Pietra, Fredrick Jelinek, John D. Lafferty, Robert L. Mercer, and Paul S. Roossin. A statistical approach to machine translation. *Computational Linguistics*, 16(2):79–85, June 1990.

[7] Peter F. Brown, Stephen A. Della Pietra, Vincent J. Della Pietra, and Robert L. Mercer. The mathematics of statistical machine translation : Parameter estimation. *Computational Linguistics*, 19(2):263–311, June 1993.

[8] G. Carroll and E. Charniak. Learning probabilistic dependency grammars from corpora. In *Working Notes, Fall Symposium Series*, pages 25–32. AAAI, 1992. Cited in [Charniak 1993].

[9] G. Carroll and E. Charniak. Two experiments on learning probabilistic dependency grammars from labeled text. In *Workshop Notes, Statistically-Based NLP Techniques*, pages 1–13. AAAI, 1992. Cited in [Charniak 1993].

[10] Eugene Charniak. *Statistical Language Learning*. MIT Press, Cambridge, MA, 1993.

[11] Noam Chomsky. Remarks on nominalization. In *Studies on Semantics in Generative Grammar*, pages 11–61. Mouton, The Hague, 1972.

[12] Noam Chomsky. *Lectures on Government and Binding*. Foris Publications, Dordrecht, Holland, 1981.

[13] Michael Cohen, Hy Murveit, Jared Bernstein, Patti Price, and Mitch Weintraub. The DE-CYPHER speech recognition system. In *Proceedings of the IEEE Interational Conference on Acoustics, Speech, and Signal Processing*, pages 77–80, 1990.

[14] S. Crespi-Reghizzi. An effective model for grammatical inference. In B. Gilchrist, editor, *Information Processing IFPI Congress 71*, New York, 1972. North-Holland.

[15] Doug Cutting, Jilian Kupiec, Jan Pedersen, and Penelope Sibun. A practical part-of-speech tagger. In *Proceedings of the Third Conference on Applied Natural Lanugage Processing*, Trento, Italy, April 1992. Association for Computational Linguistics. Also available as Xerox PARC technical report SSL-92-01.

[16] Bonnie Jean Dorr. *Machine Translation: A View from the Lexicon*. MIT Press, Cambridge, MA, 1993.

[17] Jerome A. Feldman, George Lakoff, Andreas Stolcke, and Susan Hollbach Weber. Miniature language acquisition: A touchstone for cognitive science. Technical Report TR-90-009, International Computer Science Institute, 1947 Center Street, Suite 600, Berkeley, California 94704, March 1990.

[18] William A. Gale and Kenneth W. Church. A program for aligning sentences in bilingual corpora. *Computational Linguistics*, 19(1):75–102, 1993.

[19] Fred Glover and Manuel Laguna. Tabu search. In *Modern Heuristic Techniques for Combinatorial Problems*. Blackwell Scientific Publications, 1992.

[20] Fred Glover, Eric Taillard, and Dominique de Werra. A user's guide to tabu search. unpublished monograph, 1991.

[21] E. Mark Gold. Language identification in the limit. *Information and Control*, 10:447–74, 1967.

[22] T. R. G. Green. The necessity of syntax markers: Two experiments with artificial languages. *Journal of Verbal Learning and Verbal Behavior*, 18:481–96, 1979.

[23] Joseph H. Greenberg. Some universals of grammar with particular reference to the order of meaningful elements. In Joseph H. Greenberg, editor, *Universals of Grammar*. MIT Press, Cambridge, MA, 1966.

[24] John E. Hopcroft and Jeffrey D. Ullman. *Formal Languages and Their Relation to Automata*. Addison-Wesley Publishing Company, Reading, Mass., 1969.

[25] John E. Hopcroft and Jeffrey D. Ullman. *Introduction to Automata Theory, Languages, and Computation*. Addison-Wesley Publishing Company, Reading, Mass., 1979.

[26] Ray S. Jackendoff. *X̄ Syntax : A Study of Phrase Structure*. MIT Press, Cambridge, MA, 1977.

[27] Patrick Juola, Chris Hall, and Adam Boggs. Morphological segmentation by information theory. Technical Report unassigned, Computer Science Department, University of Colorado, In preparation.

[28] Martin Kay. Algorithm schemata and data structures in syntactic processing. In Sture Allén, editor, *Text Processing: Text Analysis and Generation, Text Typology and Attribution*, pages 327–358. Almqvist and Wiksell, Stockholm, 1982.

[29] Martin Kay and Martin Roscheisen. Text-translation alignment. *Computational Linguistics*, 19(1):121–142, 1993.

[30] Edward Keenan and Bernard Comrie. Noun phrase accessibility and universal grammar. *Linguistic Inquiry*, 8:63–99, 1977.

[31] S. Kirkpatrick, C. D. Gelatt, Jr., and M. Vecchi. Optimization by simulated annealing. *Science*, 20:671–80, 1983.

[32] S.M. Lucas and R.I. Damper. Syntactic neural networks. *Connection Science*, 2(3):195–221, 1990.

[33] N. Metropolis, A. W. Rosenbluth, M. N. Rosenbluth, A.H. Teller, and E. Teller. Equations of state calculations by fast computing machines. *The Journal of Chemical Physics*, 21:1087–92, 1953.

[34] James L. Morgan, Richard P. Meier, and Elissa L. Newport. Facilitating the acquisition of syntax with cross-sentential cues to phrase structure. *Journal of Memory and Language*, 28:360–74, 1989.

[35] Kazuo Mori and Shannon D. Moeser. The role of syntax markers and semantic referents in learning an artificial language. *Journal of Verbal Learning and Verbal Behavior*, 22:701–18, 1983.

[36] Eugene W. Myers. An O(ND) difference algorithm and its variations. *Algorithmica*, 1:251–56, 1986.

[37] Makoto Nagao. A framework of a mechanical translation between Japanese and English by analogy principle. In A. Elithorn and R. Barnerji, editors, *Artificial and Human Intelligence*, pages 173–80. North-Holland, 1984.

[38] Fernando Pereira and Yves Schabes. Inside-outside reestimation from partially bracketed corpora. In *Proceedings of the Conference of 30th Annual Meeting of the Association for Computational Linguistics*, 1992.

[39] Satoshi Sato. Example-based translation approach. In *International Workshop on Fundamental Research for the Future Generation of Natural Language Processing*, Kansai Science City, Japan, July 1991. ATR Interpreting Telephony Research Laboratories.

[40] Satoshi Sato and Makoto Nagao. Toward memory-based translation. In *Proceedings of COLING-90*, volume 3, pages 247–52, 1990.

[41] C. E. Shannon. A mathematical theory of coding. In D. Slepian, editor, *Key Papers in the Development of Information Theory*, pages 5–29. IEEE Press, New York, 1948.

[42] C. E. Shannon. Prediction and entropy of printed English. In D. Slepian, editor, *Key Papers in the Development of Information Theory*, pages 42–6. IEEE Press, New York, 1951.

[43] Dan Isaac Slobin. *Psycholinguistics*. Scott, Foresman, and Company, Glenview, Ill., second edition, 1979.

[44] Dan Isaac Slobin. Crosslinguistic evidence for the language-making capacity. In Dan Isaac Slobin, editor, *The Cross-Linguistic Study of Language Acquisition*, volume 2 : Theoretical Issues, chapter 15, pages 1157–1256. Lawrence Erlbaum Associates, Inc., 365 Broadway, Hillsdale, New Jersey, 1985.

[45] Tony C. Smith and Ian H. Witten. Language inference from function words. Technical Report 1993/3, University of Waikato, New Zealand, Jan 1993.

[46] Andreas Stolcke and Stephen Omohundro. Hidden Markov model induction by Bayesian model merging. In C. L. Giles, S. J. Hanson, and J. D. Cowan, editors, *Advances in Neural Information Processing Systems V*. Morgan Kaufman, 1993.

[47] E. Sumita, H. Iida, and H. Kohyama. Translating with examples : A new approach to machine translation. In *The Third International Conference on Theoretical and Methodological Issues in Machine Translation of Natural Language*, 1990.

[48] Leonard Talmy. The relation of grammar to cognition. In Brygida Rudzka-Ostyn, editor, *Topics in Cognitive Linguistics*, pages 165–205. John Benjamins Publishing Co., Amsterdam/Philadelphia, 1988.

[49] Professor Aziz ur Rahman. *Teach Yourself Urdu in Two Months*. Azizi's Oriental Book Depot, II, K, 14/4, Nazimabad, Karachi–18, Pakistan, 22nd edition, 1958.

[50] Charles Clayton Wooters. Lexical modeling in a speaker independent speech understanding system. Technical Report TR-93-068, International Computer Science Institute, 1947 Center Street, Suite 600, Berkeley, California 94704, November 1993.

[51] Dekai Wu. Aligning a parallel English-Chinese corpus statistically with lexical criteria. In *Proceedings of the 32nd Annual Meeting of the Association for Computational Linguistics (ACL-94)*, 1994.